\documentclass[letterpaper, 10 pt, conference]{ieeeconf}
\IEEEoverridecommandlockouts
\usepackage{amsmath,amssymb,amsfonts}
\usepackage{algorithmic}
\usepackage{xcolor}
\usepackage[utf8]{inputenc}
\usepackage{textcomp}
\usepackage{graphicx}
\usepackage[version=4]{mhchem}
\usepackage{siunitx}
\usepackage{longtable,tabularx}
\usepackage{booktabs}
\usepackage{hyperref}
\usepackage{xspace}
\usepackage[caption=false]{subfig}

\makeatletter
\newcommand*{\equationname}{Eq.}
\renewcommand*{\eqref}[1]{%
	\hyperref[{#1}]{\equationname~\textup{\tagform@{\ref*{#1}}}}%
}
\makeatother

\usepackage{nomencl}
\newcommand{\nomdef}[3]{#1 (#2)\nomenclature[#3]{#2}{#1}}
\newcommand{\acrdef}[2]{\nomdef{#1}{#2}{A}}

\makenomenclature

\usepackage{etoolbox}
\renewcommand\nomgroup[1]{%
	\item[\bfseries
	      \ifstrequal{#1}{A}{Acronyms}{%
	      	\ifstrequal{#1}{S}{Symbols}{}}%
]}

\newcommand\LHCGWCONV{LHC\_GW\_CONV\xspace}

\graphicspath{{figures/}{figures/generated/}}

\newcommand{\NAtHunredPRelTime}{16.78}
\newcommand{\RelErrAtHundredPRelTime}{14.75}

\title{\LARGE \bf Enhancing Reinforcement Learning in Sensor Fusion: A Comparative Analysis of Cubature and Sampling-based Integration Methods for Rover Search Planning
}
\author{
	{Jan-Hendrik Ewers$^{1}$, Sarah Swinton$^{1}$, David Anderson$^{2}$, Euan McGookin$^{2}$, and Douglas Thomson$^{2}$}
	\thanks{This work has been submitted to the IEEE for possible publication. Copyright may be transferred without notice, after which this version may no longer be accessible.}
	\thanks{This work was supported by the Engineering and Physical Sciences Research Council, Grant/Award Number: EP/T517896/1-312561-05}
	\thanks{$^{1}$ Aerospace Sciences Research Division, University of Glasgow, Scotland 
	{\tt \small \{j.ewers.1, s.swinton.1\}@research.gla.ac.uk
	}} 
	\thanks{$^{2}$ Aerospace Sciences Research Division, University of Glasgow, Scotland 
	{\tt \small \{dave.anderson, euan.mcgookin, douglas.thomson\}@glasgow.ac.uk}
}}

\begin{document}

\maketitle

\begin{abstract}

This study investigates the computational speed and accuracy of two numerical integration methods, cubature and sampling-based, for integrating an integrand over a 2D polygon.
Using a group of rovers searching the Martian surface with a limited sensor footprint as a test bed, 
the relative error and computational time are compared as the area was subdivided to improve accuracy in the sampling-based approach. 
The results show that the sampling-based approach exhibits a $\RelErrAtHundredPRelTime\%$ deviation in relative error compared to cubature when it matches the computational performance at $100\%$. Furthermore, achieving a relative error below $1\%$ necessitates a $10000\%$ increase in relative time to calculate due to the $\mathcal{O}(N^2)$ complexity of the sampling-based method.
It is concluded that for enhancing reinforcement learning capabilities and other high iteration algorithms, the cubature method is preferred over the sampling-based method.

\end{abstract}

\section{Introduction}
\label{sect:intro}

\acrdef{Machine Learning}{ML} has gained massive popularity in the last decade
\cite{sarker_machine_2021} driven by both the ongoing research into learning algorithms and cheap computational performance
\cite{jordan_machine_2015}.
 Phenomenon such as the success of generative artificial intelligence
 \cite{stokel-walker_what_2023}
have put ML more into the eye of the public than ever before.
 However,
 behind the scenes, researchers spend a vast amount of time preparing data for learning.
 All three major forms of ML require high-quality data for the best performance: supervised learning needs well-labelled data points,
 unsupervised learning must have noise-free data for labeling,
 and reinforcement learning has to have access to a good representation of the environment.
 Studies on the impact of non-systematic noise on training have shown massively degraded performance
\cite{nettleton_study_2010,
quinlan_induction_1986}.

While significant progress has been made in developing robust ML algorithms 
\cite{guo_robust_2023, fox_taming_2017}, 
the impact of noise in reinforcement learning remains a critical challenge
\cite{sun_exploring_2023}.
In the realm of reinforcement learning noise can be beneficial.
Action noise, 
 which is the addition of randomness to the action selected by the algorithm,
 can be desirable to encourage exploration of the action space.
 Using Gaussian or Ornstein-Uhlenbeck noise has been shown to significantly improve training across popular reinforcement algorithms such as Soft Actor-Critic 
\cite{haarnoja_soft_2018}
or Deep Deterministic Policy Gradient 
\cite{lillicrap_continuous_2019}
when coupled with scheduled noise reduction
\cite{hollenstein_action_2023}.
 However,
 this scheduling shows that noise must be minimized to fine-tune the model.
 In comparison,
 observation noise in reinforcement learning,
 akin to poor-quality labelled data for supervised learning,
 is known to make training unstable 
\cite{sun_exploring_2023}.
Noise in the reward channel for reinforcement learning is particularly undesirable as it can lead to actions being misinterpreted.
\cite{everitt_reinforcement_2017} 
highlights multiple scenarios in which the agent may take advantage of a reward function leading to 
\textit{reward corruption}
and a stagnation in learning on a local maxima.
Whilst a solution to this was proposed in 
\cite{everitt_reinforcement_2017}, 
a much easier approach is to minimize the chances of this happening in the first place by implementing a robust method wth as little noise as possible.
Overall noise in ML,
when not controllable,
is undesirable as it increases the difficulty in successfully learning the policy. 

A famously noisy task is numerical integration.
 From Euler integration to complex modern algorithms,
 errors are inevitable compared to continuous methods as numerical methods are just approximations.
 However,
 reducing these errors is paramount to decreasing noise.
 It is very common to have simulation step updates within a reinforcement learning environment which require integration in multiple forms.
 This might be double integrating acceleration to get the position,
 or integrating a function within a 2D shape.
 The latter is what this research focuses on.
 This problem takes the form
\begin{equation}
I(f,H) = \int_H f(\vec x) dH
\label{eqn:integration_problem_definition}
\end{equation}
where $\vec x \in \mathbb R^2$,
 $f$ is the integrand,
 and $H$ is the collection of $s$-simplices.
 In search path planning,
 the integrand is commonly a \acrdef{Probability Distribution Map}{PDM},
 and $H$ is the path buffered by the sensor footprint $r$.
 A widely used method of integration is a sampling-based approach which calculates the integration result
 through grid sampling of the PDM and then calculating the summation of pixels within 
 $H$.
A different approach uses cubature 
\cite{lyness_survey_1994}
to achieve a similar result. 
This relies on the transformation of 
$H$
into a sum of computable simplices such as triangles or rectangles. 
\textit{cubpack++} by 
\cite{cools_algorithm_1997},
implement methods for 15 primitive regions that any simplex can be dissected into, like the aforementioned triangle and rectangle. 
The integration result is then the sum of cubature results for each simplex. 
Another approach by 
\cite{robinson_algorithm_2002}, \textit{r2d2lri}, uses a sequence of embedded lattice rules and claims to result in little computational overhead. 
Whilst 
\cite{hill_comparison_2003} found \textit{r2d2lri} to be faster during integration over a infinite domain, \textit{cubpack++} proved to be more adaptable to different scenarios without a massive speed penalty.
For the rest of this work, \textit{cubpack++} will be used for for its adaptability and adaptive properties. 

The task of numerically integrating an integrand over a 2D shape within search planning can be thought of as \textit{"How much probability has been seen by executing this mission?"} In this application, the integrand is a PDM that defines the probability of detection at a given point and the 2D shape is the search path buffered by some quantity to get a polygon. To apply reinforcement learning to this problem, the newly observed probability could be used to calculate the reward function. 
As previously outlined, this requires a fast and noiseless numerical integration method for good training characteristics. 

Search planning typically is applied in situations where exhaustive search (also known as coverage planning) is infeasible due to mission constraints such as limited resources. One such situation where a reduced vehicle lifespan is the limiting factor is searching the Martian surface for items of scientific interest based on a PDM. In 2020, NASA's Mars Mission utilized two robots: the Perseverance rover and the Ingenuity helicopter - opening the door to the use of collaborative robotic teams for planetary exploration 
\cite{farley2020}. The use of multi-rover teams for planetary exploration missions has been considered in terms of both surface 
\cite{fong2021} and cave 
\cite{vaquero2018} exploration. By using a team of planetary exploration rovers, the data collection capabilities of a mission can be increased due to the extended sensor footprint. 
Whilst this study was focused on exploratory Mars rovers, search planning has applications within other domains such as in wilderness
\cite{lin_uav_2009} or marine
\cite{furukawa_recursive_2006} search and rescue.

In this study, the two aforementioned integration methods are benchmarked against each other to determine the performance differences between them. This will inform future studies on which method is appropriate for their application. A group of rovers searching the Martian landscape with reduced sensor footprints is used as a benchmark. The generated paths are highly irregular due to collision avoidance and inter-rover communications creating a challenging 2D polygon to integrate over. The PDM for this scenario is analogous to the probability distribution of a meteor that has crashed on the surface and that the rovers are now searching for. 

The top-level mission planning and environment generation is introduced in \autoref{sect:method_mission_planning}, with modelling and path planning in \autoref{sect:method_rover_modelling}. Sampling-based and cubature integration is defined in \autoref{sect:method_sampling_integration} and \autoref{sect:method_cubature_integration} respectively. The results are presented and discussed in \autoref{sect:results} with a conclusion being drawn in \autoref{sect:conclusion}. 

\section{Method}
\label{sect:method}

\subsection{Mission Planning}
\label{sect:method_mission_planning}

The PDM is modelled as a sum of $G$ bivariate Gaussian
\cite{yao_optimal_2019} such that a point on the surface of Mars at coordinate
$\vec x \in \mathbb R^2$
has a probability of containing the meteor p
\begin{equation}
	p(\vec x) =
	\frac{1}{G} 
	\sum^G_{i=0}
	\frac
	{
		\exp{
			\left[
				-\frac
				{1}
				{2}
				(\vec x - \vec \mu_i )^T\vec\sigma_i^{-1}(\vec x - \vec \mu_i) 
			\right]
		}
	}
	{\sqrt{4\pi^2\det{\vec\sigma_i}}}
	\label{eqn:sum_of_bivariate_gaussians}
\end{equation}
where $\vec \mu_i$ and $\vec \sigma_i$ are the mean location and covariance matrix of the $i$th bivariate Gaussian respectively. For this study, these values are randomly generated with $G=4$.
\cite{lin_hierarchical_2014} used a similar method to approximate a discrete PDM through a Gaussian mixture model which employs \eqref{eqn:sum_of_bivariate_gaussians} for the final $p(\vec x)$.

This PDM is then used in conjunction with 
\LHCGWCONV
\cite{lin_uav_2009}
to generate 
$64$
mission waypoints with a search radius of
$5\si{\m}$.
The area is 
$150 \si{\m} \times 150 \si{\m}$
and is thus segmented into a 
$30 \times 30$
grid such that each cell is the size of the search radius.

\LHCGWCONV is,
at its core,
a greedy algorithm by employing the \acrdef{local hill climb}{LHC} optimization method. 
This evaluates all eight surrounding cells and moves to the highest-scoring one. 
If there is a tie, 
a 
convolution kernel $\omega$ centred around the offending cells is evaluated and the cell with the highest score 
\begin{equation}
	p^{conv}(\vec x) = \omega * p(\vec x) = 
	\sum_{i=-1}^{1}\sum_{j=-1}^{1}
	\omega(i,j)p\left(
	\vec x - 
	\begin{bmatrix}
		i \\j
	\end{bmatrix}
	\right)
\end{equation}
 is selected.
A $3 \times 3$ normalized box blur kernel is typically used and is defined as
\begin{equation}
	\omega = \frac{1}{9}
	\begin{bmatrix}
		1& 1& 1\\
		1& 1& 1\\
		1& 1& 1
	\end{bmatrix}
\end{equation}

Furthermore, 
LHC does not handle multi-modal problems well as it attempts to cover one mode completely before being able to move to the next. 
This often leads to LHC ignoring much more rewarding modes. 
To solve this, 
\acrdef{global warming}{GW} 
is introduced. 
This works by subtracting 
$C$ 
from the  PDM 
a 
$l$ 
number of times, 
where 
$C=p_{\max}/l$ and $p_{\max}$ is the global maxima. 
This then gives the updated PDM 
\begin{equation}
	p'(\vec x) = 
	\begin{cases}
		p(\vec x) - C, & p(\vec x)>C \\
		0,             & else        
	\end{cases}
\end{equation}
which the LHC uses to generate a new path. 
After all 
$l$
GW steps are evaluated,
they are scored using the original PDM 
$p(\vec x)$ 
and the path with the highest accumulated probability is selected.

Finally, each visited cell is marked a \textit{seen}, and subsequent evaluations of $p(\vec x)$ at this cell will return $0$. 
This discourages revisiting a position but does not entirely forbid it.

\subsection{Rover Modelling and Path Planning}
\label{sect:method_rover_modelling}

\subsubsection{Mars Map Generation}
\begin{figure}[htbp]
    \centering
    \includegraphics[
		clip, trim=2cm 7.5cm 1cm 8cm, width=0.75\linewidth
	]{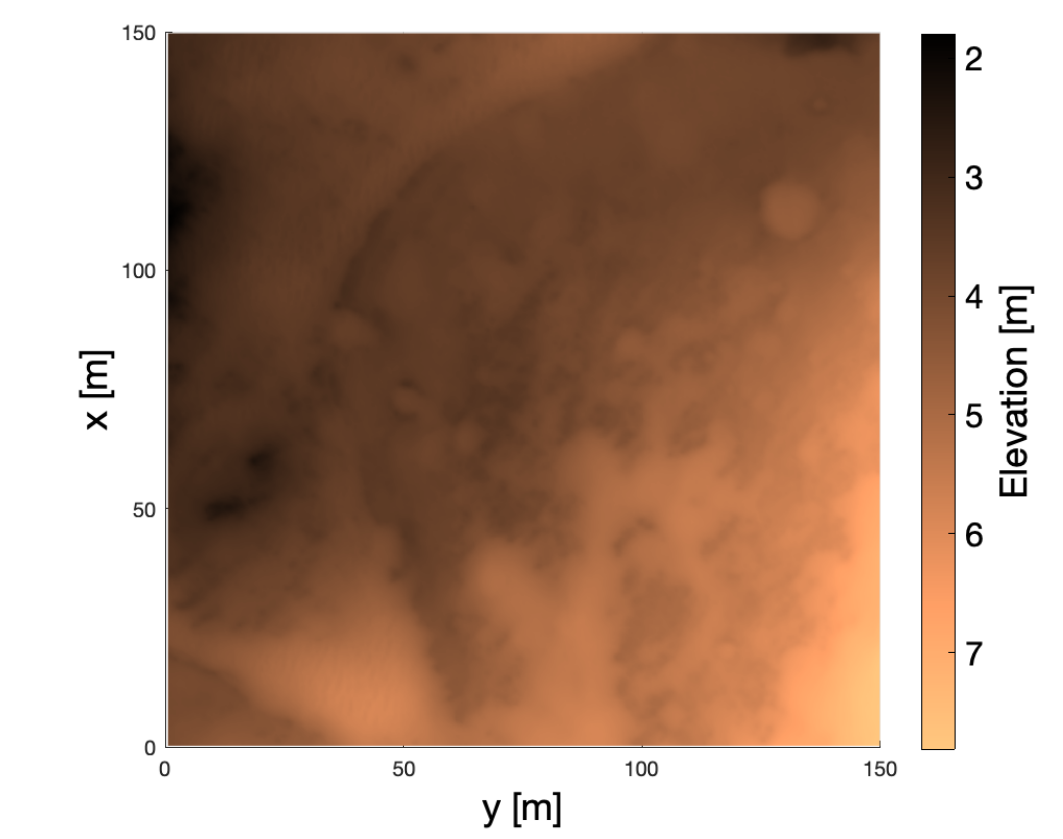}
    \caption{3D terrain model of the selected mission site} \label{fig:3dTerrainModel}
\end{figure}

A 3D terrain model has been created using data from the High-Resolution Imaging Science Experiment
(HiRISE) on the Mars Reconnaissance Orbiter. The map is a matrix of latitude, longitude, and
elevation data, and has a resolution of $0.3\si\meter$ per pixel
\cite{bergstrom2004}. For this
work, a mission site has been selected from within the Jezero crater. The generated mission site
consists of a $600 \times 600$ block grid. \autoref{fig:3dTerrainModel} shows the 3D terrain model that has been generated, based on previous work at the University of Glasgow
\cite{baxter2022}.

\begin{figure}[htbp]
    \centering
    \includegraphics[clip, trim=1cm 7.5cm 1cm 8cm, width=0.85\linewidth]{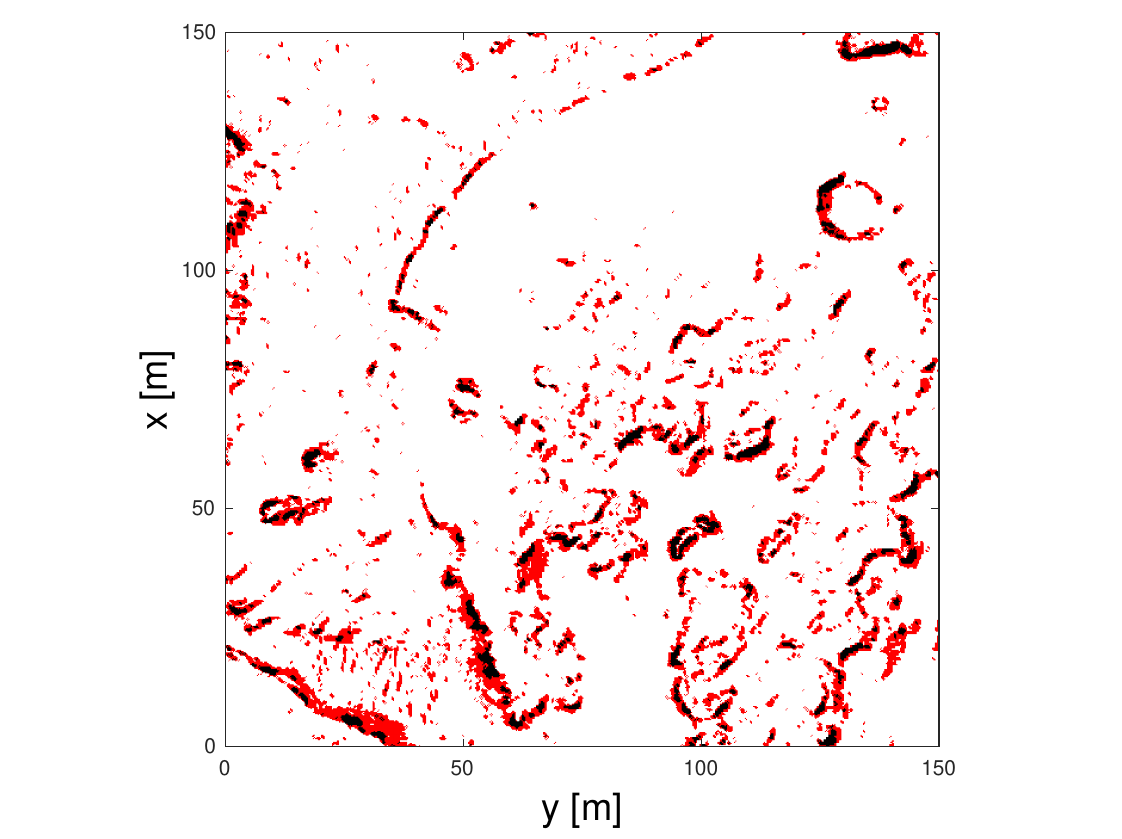}
    \caption{Traversability analysis of the selected mission site. Traversable terrain is shown in white, high-risk terrain is shown in red, and impassable terrain is shown in black.} \label{fig:travAnalysis}
\end{figure}

The 3D terrain model is analysed in terms of the elevation of neighbouring blocking in the grid. A given block inherits the worst-case slope angle, $\theta$, from its eight neighbouring blocks. Three traversability categories have been defined: traversable 
($\theta < 10^{\circ}$), 
high-risk 
($10^{\circ} 	\leq \theta \leq 15^{\circ}$), 
and impassable 
($\theta > 15^{\circ}$).
The thresholds of the three traversability categories have been set in line with the nominal operational limits of the Perseverance rover
\cite{rankin2020}. \autoref{fig:travAnalysis} shows the resulting traversability map for the selected mission site.
\subsubsection{Rover Model}
The robot utilised for analysis and testing in this work is the \acrdef{rocker bogie runt}{RBR}. This rover has a small form factor ($0.271\si\meter \times 0.251\si\meter \times 0.144\si\meter$), and a six-wheel rocker bogie suspension in line with the baseline mobility characteristics of current planetary exploration robotos
\cite{flessa2014}. The RBR has six wheels; three wheels on each side, where a front wheel is connected to the rocker, and the middle and rear wheels are connected to the bogie. Each of the six wheels is driven by its own 6V DC motor. Unlike NASA's Mars rovers, which utilise steerable wheels, all RBR wheels are fixed and cannot rotate. Therefore, the differential drive steering method is used. For this work, a team of five rovers is utilised, each with a search footprint diameter of $d_{rover}=1\si\meter$. 

\subsubsection{Generating Safe Paths}
The paths of each member of the rover team must be coordinated such that no collisions occur between them as they traverse paths toward their respective targets. As planetary exploration rovers operate in such remote and hazardous environments, collisions could cause the loss of the rovers involved and severe degradation to the group's data collection capabilities.

For each target point generated by \LHCGWCONV,
a set of safe paths is generated using prioritised planning coordination
\cite{swinton_novel_2022}.
Each rover path is given an arbitrary priority number. A path planning attempt is then made for the first rover, using RRT*. A simulation of the rover is run to obtain 4D pose data. The path of the second rover the then planned, and its behaviour is simulated. The 4D pose data for both rovers is compared to check for collisions. If a collision occurs, the path of the lower priority rover is re-planned until a safe path is found. If no collisions occur, the path is saved and the algorithm moves to the next rover. This process continues until all five rovers have safe paths.

\subsection{Path Analysis}

Each rover generates a unique path as it executes the mission which is buffered by the search footprint diameter $d_{rover}$. The total search area $H_{total}$ is then the union of the set of buffered paths $H$ such that 
\begin{equation}
	H_{total} = \bigcup_{h \in H} h
\end{equation}

This can be extended to be a function of time $t$ by
\begin{equation}
	H_{total}(t) = \bigcup_{h \in H(t)} h
\end{equation}
where $H(t)$ is the set of buffered paths covered by the rovers up until $t \si{\second}$. If $t_{\max}\si{\second}$ is the time taken for the entire mission to be completed, it then follows that $H_{total} = H_{total}(t_{\max})$ and $H_{total}\neq H_{total}(t)~\forall~t  < t_{\max}$.

\subsubsection{Sampling-based Integration}
\label{sect:method_sampling_integration}

Accessing the value of $p(x,y)$ at a given coordinate is trivial, and is as computationally expensive as the definition of $p(x,y)$ itself. A harder task is determining if the point at $(x,y)$ is within $H_{total}(t)$. 

\begin{figure}[htbp]
	\centering
	\includegraphics[width=0.7\linewidth]{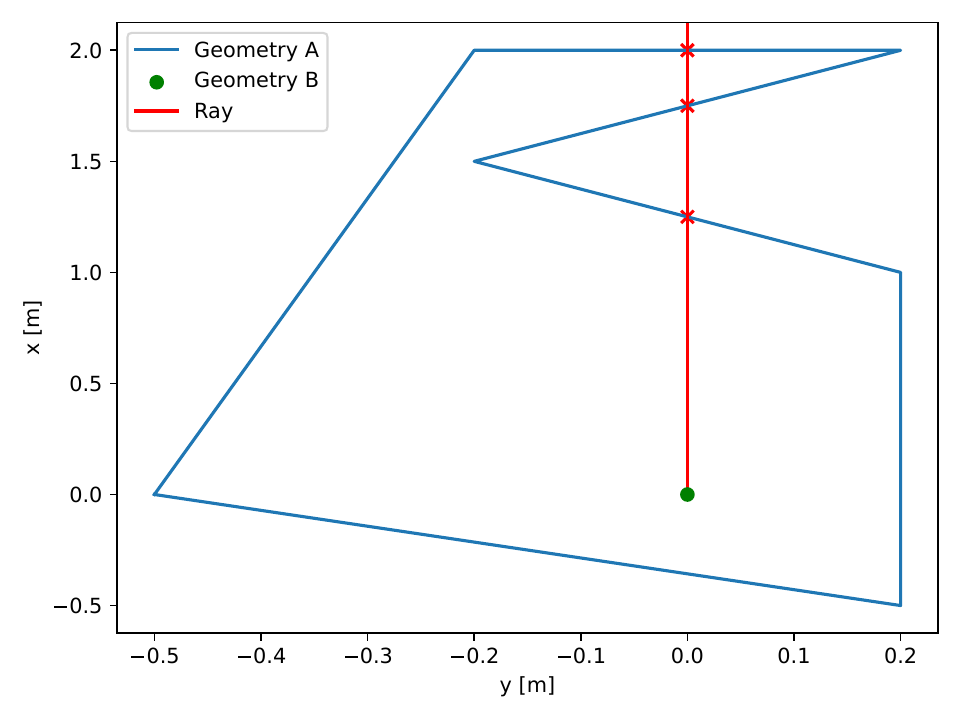}
	\caption{Ray casting method to check if a point (geometry A) is within a non-convex polygon (geometry B). If the ray originating from A intersects B an odd amount of times (three, in this example), then A is within B.}
	\label{fig:method_within_raycasting_example}
\end{figure}

A popular method is ray casting along a straight line from a point
\cite{sutherland_characterization_1974}. If this ray intersects $H_{total}(t)$ (geometry B in the example seen in \autoref{fig:method_within_raycasting_example}) an odd amount of times then the point is contained by the buffered search paths. 

Another popular method uses the 
\acrdef{Dimensionally Extended Nine Intersection Matrix}{DE-9IM}\cite{clementini_small_1993}. 
This is calculated using
\newcommand{\dimcap}[2]{D(A^#1 \cap B^#2)}
\begin{multline}
	DE-9IM(A,B) = \\
	\begin{bmatrix}
		\dimcap{I}{I} & \dimcap{I}{B} & \dimcap{I}{E} \\
		\dimcap{B}{I} & \dimcap{B}{B} & \dimcap{B}{E} \\
		\dimcap{E}{I} & \dimcap{E}{B} & \dimcap{E}{E} 
	\end{bmatrix}
\end{multline}

where $D(x)$ is the dimension of 
$x$ 
($D(x) \in (0,1,2)$
if 
$x \neq \emptyset$, 
and $D(x) = -1$ if $x = \emptyset$), 
and the superscripts $I$, $B$, and $E$ denote the interior, boundary, and exterior regions respectively.
This matrix can be used to calculate $47$ different predicates for $A$ and $B$ to be in. The predicate of interest here is the topological \textit{within}, which is described by the pattern matrix
\begin{equation}
	within
	=
	\begin{bmatrix}
		T & * & F \\
		* & * & F \\
		* & * & * 
	\end{bmatrix}
\end{equation}
where an entry in the matrix is marked $T$ if $D(x) \in (0,1,2)$, $F$ if $D(x) = -1$, and $*$ if $D(x) \in (-1,0,1,2)$. 

The DE-9IM approach is $10.05\%$ faster than the ray-casting method when dealing with the high-vertex polygons generated by the buffered paths $H$
and is the method used for the rest of this study.

As touched on in \autoref{sect:intro}, the sampling method is the sum of pixels within the polygon written as
\begin{equation}
	I(f,H) =
	\frac{A}{NM}
	\sum^N_{n=0}
	\sum^M_{n=0}
	\begin{cases}
		0,
		  & (n,  
		m)~\text{not within}~H \\
		f(n,
		m),
		  & else 
	\end{cases}
	\label{eqn:sampling_based_integration}
\end{equation}
where $N$ and $M$ are the dimensions of the rasterisation,
$A$ is the area enclosed by the rasterisation, $f(n,
m)$ is the integrand (akin to $f(\vec x)$ from \eqref{eqn:integration_problem_definition}) which is a function accessing the pixel value at $(n,
m)$, and $H$ is a single polygon.
This method can be fast with low values of $N$ and $M$,
and accurate with high values alike.
However,
this method highly suffers from aliasing around the borders of $H$.
Methods to get around the aliasing problem have been tried 
\cite{ewers_optimal_2023}
with major performance penalties.
Nonetheless,
due to its simplicity, it is prominently used in the literature 
\cite{lin_hierarchical_2014,
	morin_ant_2023,
	subramanian_probabilistic_2020,
	ebrahimi_autonomous_2021,
hong_uav_2011}.

\subsubsection{Cubature Integration}
\label{sect:method_cubature_integration}

In order to convert the simplex $H$ into something usable for \textit{cubpack++}, it must first be subdivided into a set of the 15 primitive simplices, 
with the simplest and most ubiquitous of these being the triangle. 
As $H$ is likely to have many holes and unlikely to be convex, 
the constrained Delaunay triangulation
\cite{chew_constrained_1987}
is used to get the set of triangles $T$. 

\begin{figure}[htbp]
	\centering
	\includegraphics[width=0.4\linewidth]{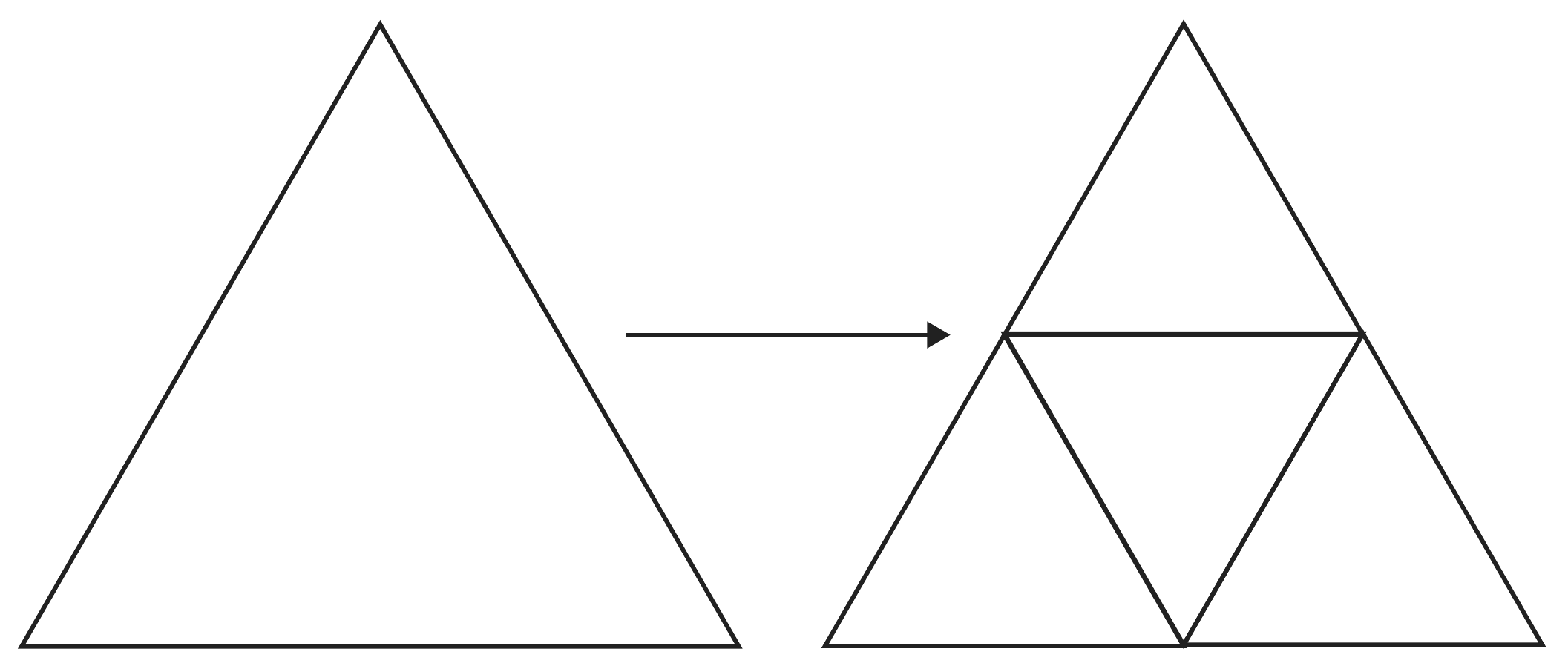}
	\caption{Subdivision of the unit triangle into four smaller triangles.}
	\label{fig:dcutri_triangle_division}
\end{figure}

At the beginning of the algorithm, 
the integration estimate 
$\hat q$ 
and error estimate 
$\hat e$ 
is evaluated for every triangle in $T$ where the unit triangle is transformed to the given triangle, 
while the cubature formula maintains its given degree. 
To calculate 
$\hat q$, 
the cubature formula of polynomial degree 13 with 37 points 
\cite{wandzurat_symmetric_2003} 
is used. 
$\hat e$ 
is calculated using several null rules as defined in
\cite{berntsen_algorithm_1992}.
If 
$\hat E = \sum \hat e > max(\epsilon_a, \epsilon_r | \hat Q |)$ 
(where 
$\epsilon_a, \epsilon_r$ 
are user-defined absolute and relative error tolerances respectively) 
is above the tolerance set, 
the triangle with the highest error is subdivided into four smaller triangles as seen in 
\autoref{fig:dcutri_triangle_division}. 
Each new triangle has its integral and error estimates evaluated, and the four replace the original triangle.
This process is continued until either 
$\hat E$
is below the tolerance, or a maximum number of calculations has been exceeded.

\section{Results}
\label{sect:results}

\begin{figure}[htbp]
	\centering
	\includegraphics[width=0.99\linewidth]{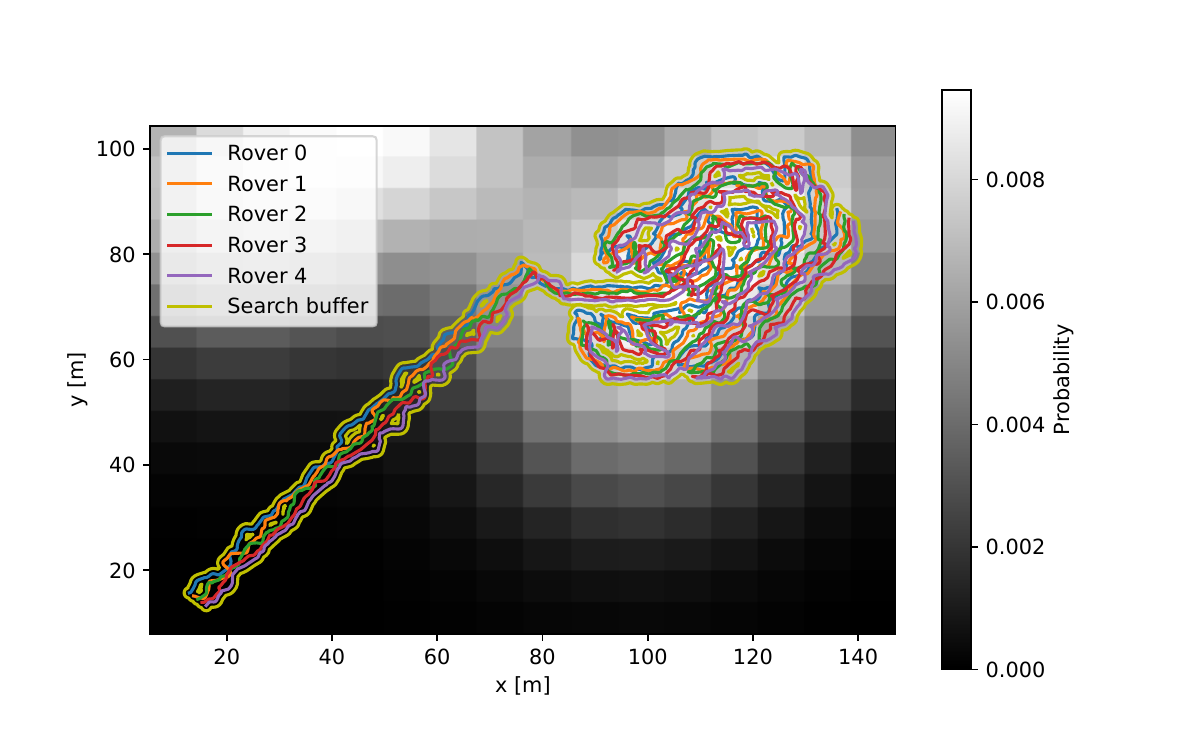} 
	\caption{Rover trajectories over the area given the mission planned by \LHCGWCONV with a random PDM $p(\vec x)$. $N=16$ was selected to emphasize the issue with sampling-based integration methods.}
	\label{fig:rover_path_over_pdm}
\end{figure}

As discussed in \autoref{sect:method_mission_planning}, LHC is a greedy algorithm that prioritizes local modes. This can be seen in \autoref{fig:rover_path_over_pdm} at $(80,80)\si{\meter}$ where the mission abruptly leaves the saddle between the modes to ascend the local probability mode. The rovers attempt to avoid collision and obstacles, creating small holes in the buffered search path as well as a highly irregular exterior. This creates a challenging benchmark to compare and contrast the methods.

To benchmark the methods, the cubature method is evaluated for each PDM and corresponding paths and the mean of this result (time and integration result) was deemed the baseline. Overall, $105$ missions have been simulated using the mission planning from \autoref{sect:method_mission_planning} and the rover simulation from \autoref{sect:method_rover_modelling}. In this case, the sampling method is evaluated from $N=10$ to $N=300$, and the relative error and relative time are calculated with
\begin{equation}
	e_{rel} = \frac{|c-s|}{c}
\end{equation}
where $c$ and $s$ are the cubature and sampling values respectively. 

\begin{figure}[htbp]
	\centering
	\includegraphics[width=0.8\linewidth]{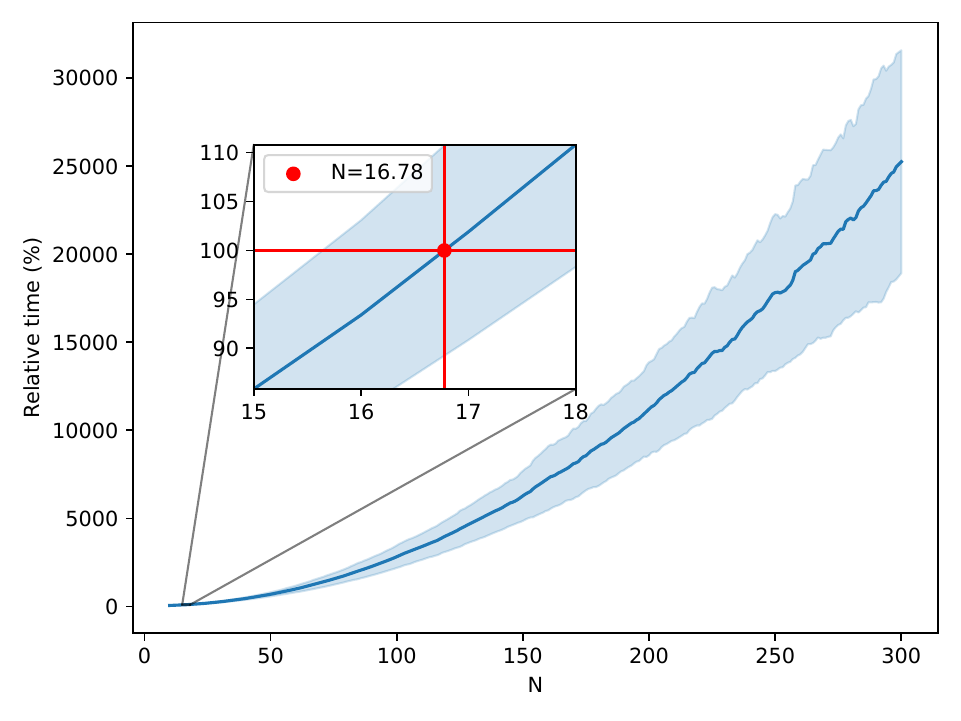} 
	\caption{The relative time as $N$ increases showing a trend towards an exponential increase as $N \rightarrow \infty$.}
	\label{fig:benchmark_relative_time_over_N}
\end{figure}

From \eqref{eqn:sampling_based_integration}, it can be seen that the grid that the sampling method uses grows in $\mathcal{O}(NM)$. For the remainder of this study, $M=N$ for simplicity making this algorithm $\mathcal{O}(N^2)$. This is evident from \autoref{fig:benchmark_relative_time_over_N} where it can be seen that after $N=\NAtHunredPRelTime$, the cubature method is as fast to compute as the sampling method. After $N=300$, this value is at $25000\%$. For reinforcement learning, where billions of steps are taken in the modelled environments during training, this can quickly add days to the learning time. 

\begin{figure}[htbp]
	\centering
	\includegraphics[width=0.8\linewidth]{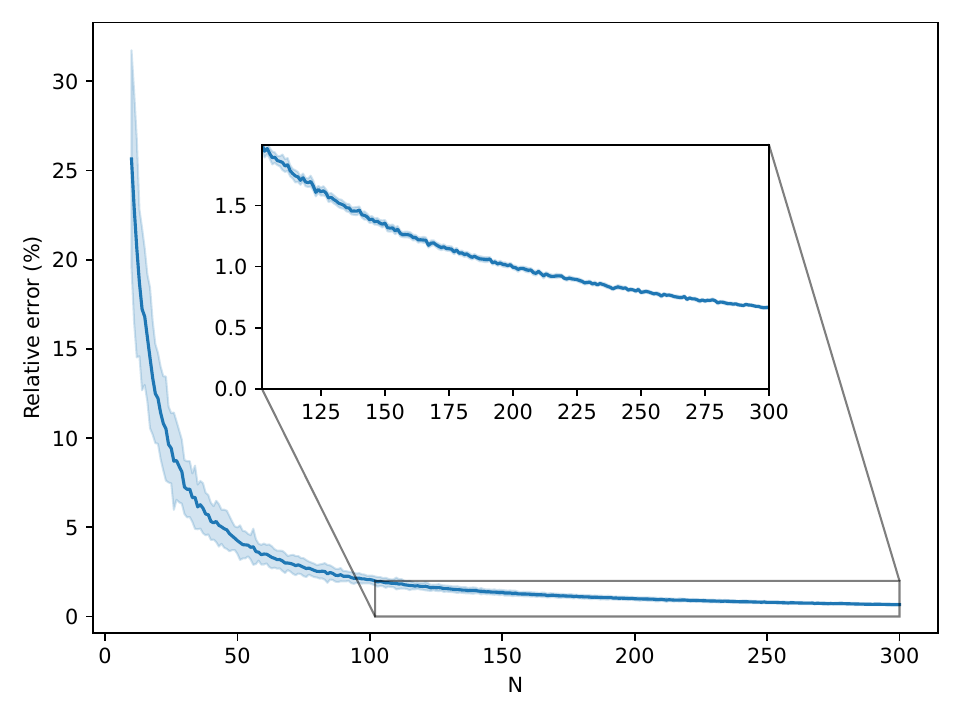} 
	\caption{The relative error as $N$ increases showing a trend towards $0\%$ as $N \rightarrow \infty$.}
	\label{fig:benchmark_relative_error_over_N}
\end{figure}

\autoref{fig:benchmark_relative_error_over_N} shows the exponential decay of the relative error as $N$ increases. At the crossing point of $N=\NAtHunredPRelTime$, the relative error is $\RelErrAtHundredPRelTime\%$. Depending on the application, this level of accuracy might be deemed sufficient. However, for reinforcement learning, as outlined in \autoref{sect:intro}, any noise is to be mitigated as it can lead to unstable training and therefore diminished results. Even at the high value of $N=300$, the relative error is still above $0.5\%$. This size of error is likely to be acceptable, however the computational cost is too high to justify this method.

A smaller $N$ might lead to faster computation, the physical meaning of $N$ must be considered. The individual rovers have a sensor diameter of $1\si\meter$ and the area has dimensions $150\si\meter \times 150 \si\meter$. To ensure that each grid cell is the same size as that of a single rover's sensor, $N = \frac{150}{1^2} = 150$ which would lead to a $800\%$ relative time difference between the two methods.

\begin{figure}[htbp]
	\centering
	\includegraphics[width=0.8\linewidth]{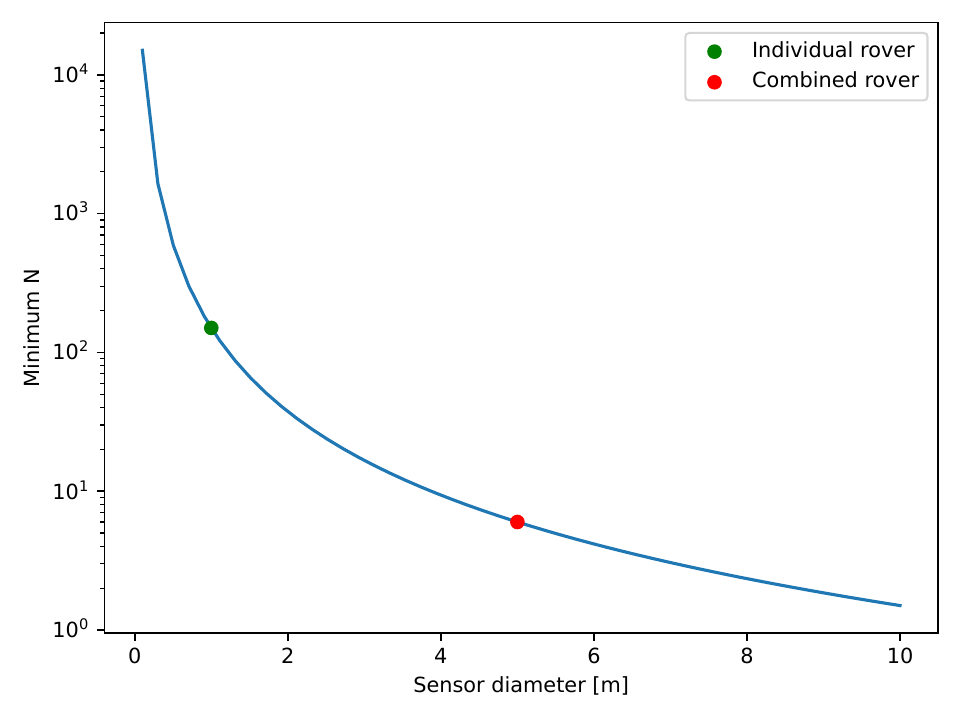} 
	\caption{Minimum $N$ required as a function of the sensor diameter.}
	\label{fig:sensor_size_to_N}
\end{figure}

The impact of the varying maximum number of coordinates per trial, denoted as $N_{coords}$, on the relative error was examined to ensure it was not a dependent variable. 
The null hypothesis posited that $N_{coords}$ does not affect the relative error. 
Following the application of an ordinary least squares, 
a $p$-value of $0.1404$ was computed surpassing the level of significance ($0.05$).
Thus, 
the null hypothesis cannot be rejected indicating that there is no significant evidence to suggest that 
$N_{coords}$ influences the relative error. 

\section{Conclusion}
\label{sect:conclusion}

This study presents a comprehensive analysis comparing cubature and sampling-based integration of an integrand over a 2D polygon. 
An example scenario requiring the search of an area on Mars by a coordinated group of rovers was used as a test bed.
These rovers completed a search mission created from the integrand; 
a random PDM.
The union of the resultant buffered paths was then used as the 2D polygon. 
Using the trajectories from $105$ missions, 
the aggregated data analysis showed that the cubature approach is substantially more accurate and significantly faster. 
Both of these qualities are important within reinforcement learning, 
where marginal performance gains can quickly snowball into days of saved training time. 
Furthermore, 
these results also translate to embedded devices where the computational budget is typically limited. 
This could be the case for the likes of onboard navigation systems on a self-sufficient rover similar to the one used in this research. 

Overall, the results of this study highlight the superiority of the cubature approach in terms of accuracy and computational efficiency, 
particularly in scenarios requiring coordinated search missions like those encountered in reinforcement learning and embedded systems for autonomous navigation.
However, 
it's important to acknowledge the limitations of this work. 
While the focus was on comparing two prevalent integration methods, 
numerous other techniques warrant exploration. 
As highlighted in \autoref{sect:intro}, 
other potentially faster but less adaptable cubature algorithms exist which could result in an even great performance increase. 
Additionally, 
the requirement of a continuous integrand for the cubature method may pose challenges in certain scenarios, 
though potential solutions using approximations have been discussed. 
Furthermore, it is important to note that the influence of problem characteristics (deterministic or stochastic, complexity, etc.) on the comparison remains unexplored.
By recognizing these limitations, 
future research can build upon these findings to further enhance integration techniques for a wide range of applications.

\bibliography{references.bib,references.sarah.bib}

\end{document}